\documentclass[letterpaper]{article} 
\usepackage[submission]{aaai23}  
\usepackage{times}  
\usepackage{helvet}  
\usepackage{courier}  
\usepackage[hyphens]{url}  
\usepackage{graphicx} 
\usepackage{natbib}  
\usepackage{caption} 
\usepackage{algorithm}
\usepackage{algorithmic}
\usepackage{dsfont}
\usepackage{booktabs}
\usepackage{multirow}
\usepackage{makecell}
\usepackage{xspace}
\usepackage{xcolor}
\usepackage{amsmath}
\usepackage{amssymb}
\usepackage{enumitem}
\usepackage{subfigure}

\usepackage{newfloat}
\usepackage{listings}
\DeclareCaptionStyle{ruled}{labelfont=normalfont,labelsep=colon,strut=off} 
\floatstyle{ruled}
\newfloat{listing}{tb}{lst}{}
\floatname{listing}{Listing}
\makeatletter
\DeclareRobustCommand\onedot{\futurelet\@let@token\@onedot}
\def\@onedot{\ifx\@let@token.\else.\null\fi\xspace}

\renewcommand*{\@fnsymbol}[1]{\ensuremath{\ifcase#1\or \dagger\or *\or \ddagger\or
   \mathsection\or \mathparagraph\or \|\or **\or \dagger\dagger
   \or \ddagger\ddagger \else\@ctrerr\fi}}

\makeatother

\title{FairCLIP: Social Bias Elimination based on Attribute Prototype Learning and Representation Neutralization}

\author{
Junyang Wang \and Yi Zhang \and Jitao Sang \thanks{Corresponding author}
\affiliations
School of Computer and Information Technology \& Beijing Key Lab of Traffic Data Analysis and Mining, Beijing Jiaotong University\\
\emails
\{junyangwang, yi.zhang, jtsang\}@bjtu.edu.cn}

\usepackage{bibentry}

\begin{document}
\maketitle

\begin{abstract}
The Vision-Language Pre-training (VLP) models like CLIP have gained popularity in recent years. However, many works found that the social biases hidden in CLIP easily manifest in downstream tasks, especially in image retrieval, which can have harmful effects on human society. In this work, we propose FairCLIP to eliminate the social bias in CLIP-based image retrieval without damaging the retrieval performance achieving the compatibility between the debiasing effect and the retrieval performance. FairCLIP is divided into two steps: Attribute Prototype Learning (APL) and Representation Neutralization (RN). In the first step, we extract the concepts needed for debiasing in CLIP. We use the query with learnable word vector prefixes as the extraction structure. In the second step, we first divide the attributes into target and bias attributes. By analysis, we find that both attributes have an impact on the bias. Therefore, we try to eliminate the bias by using Re-Representation Matrix (RRM) to achieve the neutralization of the representation. We compare the debiasing effect and retrieval performance with other methods, and experiments demonstrate that FairCLIP can achieve the best compatibility. Although FairCLIP is used to eliminate bias in image retrieval, it achieves the neutralization of the representation which is common to all CLIP downstream tasks. 
This means that FairCLIP can be applied as a general debiasing method for other fairness issues related to CLIP.
\end{abstract}

\begin{figure}
\centering
\includegraphics[width=0.44 \textwidth]{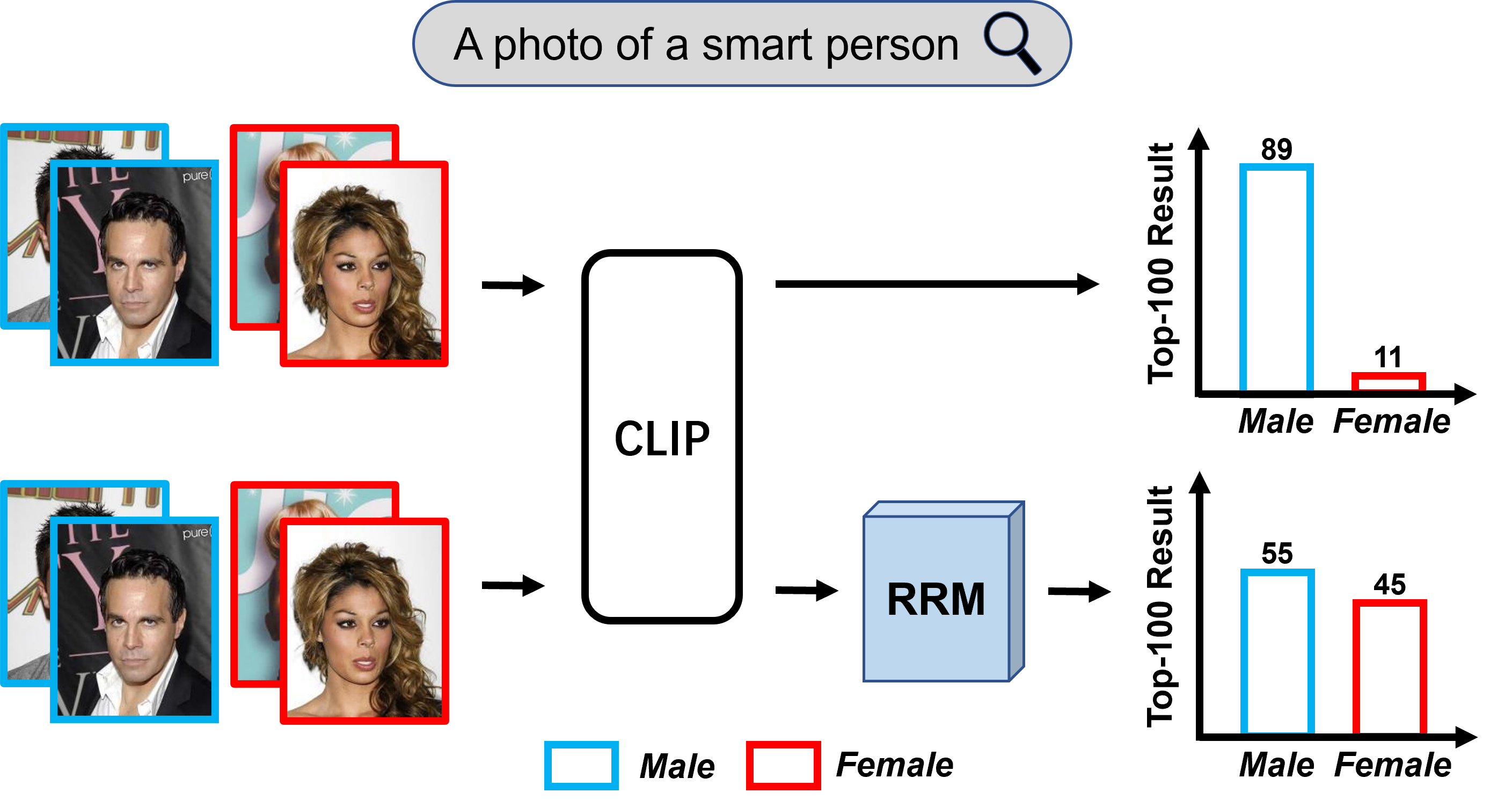}
\caption{\label{fig:intro}Example of the bias in CLIP-based image retrieval. When retrieving the face dataset using the specific text with no gender hint, almost all of the first returned 100 samples are male (top). We use Re-Representation Matrix (RRM) to neutralize the visual representations to achieve debiasing. After debiasing, the number of male and female samples is relatively balanced (bottom).}
\end{figure}

\section{Introduction}

The past few years have witnessed the rapid development of Vision-Language Pre-training (VLP) models \cite{chen2020uniter, cui2021rosita, li2020oscar, zhang2020devlbert}. Among them, OpenAI's CLIP \cite{radford2021learning} stands out. CLIP has strong zero-shot performance and can be used directly for many downstream tasks without fine-tuning. One typical task is image retrieval. With the help of aligned image-text semantic space from CLIP, it is possible to retrieve images by text. Although image retrieval is a relatively benign downstream task \cite{berg2022prompt}, \cite{geyik2019fairness} still considered that image retrieval exists fairness issues. For example, as shown in Figure \ref{fig:intro} (top), when using CLIP as the retrieval model and ``A photo of a smart person'' as the retrieval text to retrieve images on the face dataset, 89 of the first 100 samples in the result are male, even though there is no explicit gender hint.

Prioritizing fairness is of central importance in artificial intelligence (AI) systems, however, there exists very limited work addressing the bias in CLIP-based image retrieval. \cite{berg2022prompt} used an adversarial learning method for debiasing. They added learnable word vectors before the retrieval text to interfere with CLIP to capture gender information from the text. \cite{wang2021gender} used a dropout-like editing method. They dropped the dimensions that are highly correlated with gender in the visual and text representation. The above works failed to consider the multimodal property of CLIP. Since their methods are migrated from unimodal bias work, they ignore the fact that biases in the VLP model are generated from the interaction between the two modalities, making it difficult to guarantee the alignment of the two modalities after debiasing. A strong debiasing can destroy the semantic space while a weak debiasing has a very limited effect. This results in the incompatibility between the debiasing effect and retrieval performance.

To better consider the multimodal property of CLIP for debiasing, there are two challenges to be addressed. First, to achieve the debiasing for specific attributes, we need to model the concepts of attributes in CLIP. A natural idea is using the manual query to extract the concepts. However, the manual query usually generalizes poorly to downstream datasets \cite{zhou2022conditional}. This can lead to a deviated debiasing goal, thus reducing the debiasing effectiveness. Second, it is impractical to retrain CLIP because of the significant difference between pre-training tasks and downstream tasks. This means that the methods based on retraining are difficult to migrate to CLIP.

\begin{figure*}
\centering
\includegraphics[width=0.94 \textwidth]{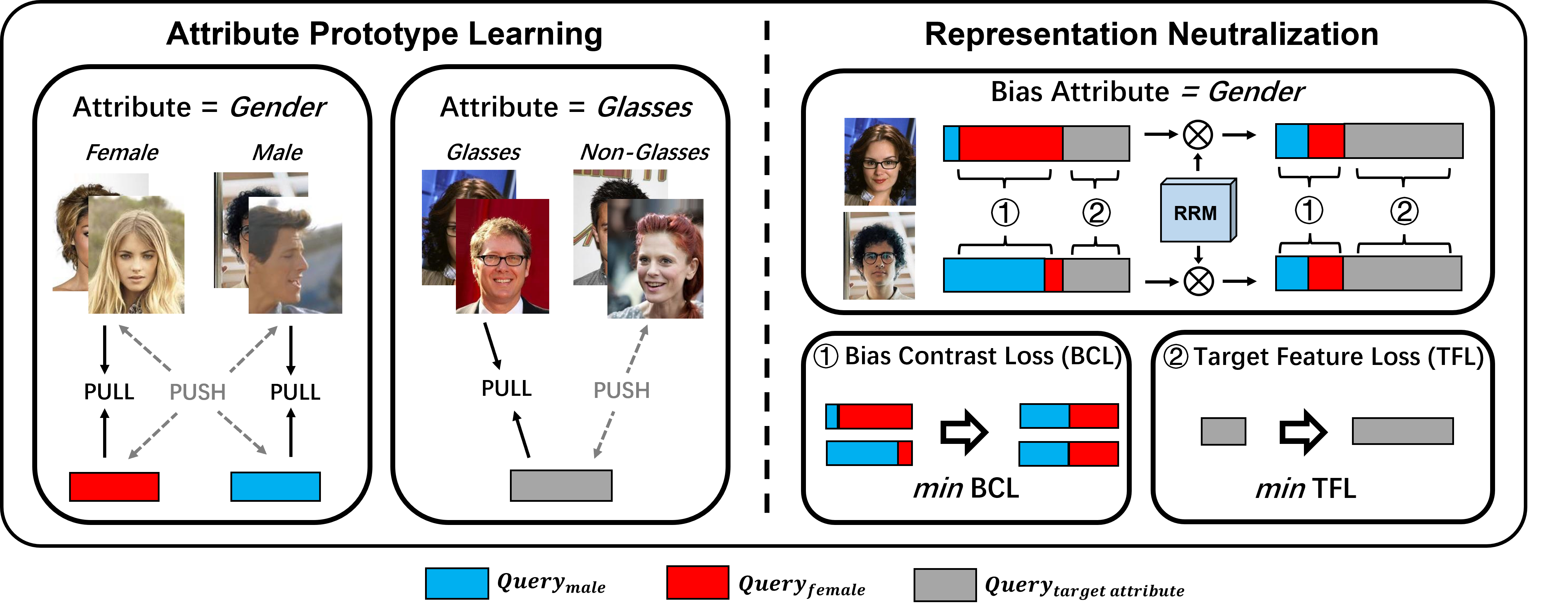}
\caption{\label{fig:method}The schematic of FairCLIP. FairCLIP is divided into two steps: Attribute Prototype Learning (left) and Representation Neutralization (right). We use glasses as one of all target attributes to show.}
\end{figure*}

Inspired by optimization methods in CLIP \cite{gao2021clip, zhou2022conditional}, we propose FairCLIP as the Figure \ref{fig:method} to achieve the debiasing. The FairCLIP is divided into two steps: (1) Attribute Prototype Learning (APL). To more accurately extract the concept of attributes that match the distribution of downstream datasets, we use the structure of a query with learnable word vectors prefixes; (2) Representation Neutralization (RN). We use the Re-Representation Matrix (RRM) which is a square matrix with the same dimensions as the representation layer of CLIP to achieve the neutralization of representation. In this step, we first analyze the bias in image retrieval and divide the attributes into target and bias attributes. The bias stems from two parts. First, the divergence in the representation of bias attributes can directly cause divergence in the representation of groups with different bias attributes. Second, the bias attributes are often more significant compared to the target attributes, which makes the bias attributes have a greater impact on the retrieval results. For target and bias attributes, we set the training constraints of RRM separately.

We summarize the contributions as follows:
\begin{itemize}
\item We propose Attribute Prototype Learning (APL) that models the concepts of target and bias attributes in CLIP. To take advantage of the multimodal property of CLIP, we use the query with learnable word vector prefixes for the attribute concepts extraction.
\item We used the extracted concepts to analyze that bias is affected by both target and bias attributes. To eliminate the bias, we set the training constraints of the Re-Representation Matrix (RRM) for target and bias attributes separately to achieve Representation Neutralization (RN).
\item We examine the debiasing effect on face datasets and examine retrieval performance on image-text retrieval datasets. We compare the results with other debiasing methods. The experiments demonstrate that FairCLIP not only achieves the best debiasing effect but also has little degradation in retrieval performance.
\end{itemize}

\section{Background and Related Work}

\subsection{CLIP}

Without fine-tuning, a variety of downstream tasks can be achieved by CLIP. CLIP encodes images and text separately and calculates the similarities between them. The similarities can be used for tasks such as classification, retrieval, etc. The performance of CLIP can be close to or even better than fine-tune models \cite{radford2021learning}. Much research work has been done on applying CLIP to application scenarios such as image segmentation \cite{xu2021simple}, caption generation \cite{dai2022enabling, mokady2021clipcap}, image generation \cite{wang2022clip}, and target detection \cite{du2022learning}. \cite{shen2021much} applied CLIP's modules to other VLP models to improve the performance. Many developers have developed applications based on CLIP for different purposes in spaces of Hugging Face\footnote{https://huggingface.co/spaces}. What can be confirmed is that CLIP is making multimodal work more efficient.

\subsection{Fairness in CLIP and Image Retrieval}

Beyond performance, fairness plays a critical role in the trustworthy deployment of VLP models and is highly endorsed by many VLP model designers \cite{li2021align}. Although CLIP has been widely used, \cite{schuhmann2021laion} argued that lacking the constraint of high-quality fine-tuning data, the model can only depend on the pre-training data that often suffers from human social biases. These biases are highly susceptible to being captured by the model during the pre-training phase \cite{steed2021image}. Several works have focused on the fairness issue in CLIP. \cite{agarwal2021evaluating} found that CLIP is more likely to classify blacks as bad. \cite{dehouche2021implicit} found that CLIP can extract gender information from neutral text. \cite{wang2021assessing} found that different languages in CLIP show different biases. \cite{wolfe2022evidence} demonstrated that CLIP learns stereotypes about gender and race.

The biases in CLIP have the potential to manifest in image retrieval, which can cause serious social implications \cite{geyik2019fairness}. But limited work addresses this issue. \cite{berg2022prompt} found that the results of image retrieval in face datasets exist gender bias when using a certain neutral query. They added a learnable prefix before the query to eliminate the gender information and used adversarial training to optimize it. \cite{wang2021gender} found that gender bias also exists when retrieving images in datasets such as MSCOCO and Flickr. They used a post-processing method called CLIP-clip. The idea is to discard the dimensions of concept representation that are most relevant to gender. In the above works, the ideas of debiasing are just a simple migration of the methods in unimodal bias elimination. They failed to focus on the multimodal property of CLIP in the process of debiasing, which results in the incompatibility of the debiasing effect and retrieval performance. 

Two challenges need to be addressed to better use the multimodal property of CLIP for debiasing. First, the concept of a visual attribute is difficult to extract directly from the visual modality in CLIP. A natural idea is to use a manual query that describes this visual attribute. However, the manual query often can’t generalize well to downstream datasets so it often cannot achieve the best result \cite{zhou2022conditional}. Second, CLIP can’t be retrained. This is because CLIP uses a large amount of image-text data to construct a large-scale similarity matrix for comparison learning in the pre-training phase. Such a pre-training task is almost impossible to reproduce, and thus any optimization would damage the semantic space of CLIP. This means that many traditional fairness methods that are based on retraining are difficult to migrate to CLIP. To address the above two challenges, we propose the debiasing method FairCLIP.

\section{Methodology}

We denote the images in the dataset by $I={I_1, I_2, \ldots, I_n}$, where $I_i$ denotes the $i_{th}$ image. We use $T_B$ to denote the retrieval text. In CLIP, the visual and text encoders are two separate structures. We denote the visual encoder by $F_V$ and the text encoder by $F_L$. The visual representation $V$ and text representation $L$ are shown by the following equation:

\begin{equation}
V=\{V_1,V_2,\ldots,V_n\}=\{F_V(I_1),F_V(I_2),\ldots,F_V(I_n)\}
\end{equation}
\begin{equation}
L=F_L(T_B)
\end{equation}

After getting the representation, we calculate the similarity between $V$ and $L$, and then get the similarity set $S$:

\begin{equation}
S=Similarity(V,L)=\{S_{i,L} \mid i=1,2,……,n\}
\end{equation}
\begin{equation}
S_{i,L}=\cos<V_i,L>=\frac{V_i}{\|V_i\|}\cdot\frac{L}{\|L\|}
\end{equation}
where $S_{i,L}$ is the cosine similarity between the $i_{th}$ visual representation in $V$ and $L$. Then, we sort $S$ by similarity. Finally, the set of top $k$ images is the retrieval result $R_k$.

\subsection{Attribute Prototype Learning}

In the visual modality, the concept extracted by the visual encoder is the representation of the overall concepts of an image. Since the concepts of different attributes is difficult to decouple, the effectiveness of methods like \cite{wang2021gender}, which target a certain attribute in the image representation for debiasing, is greatly reduced. For a visual attribute, we would like to use the cross-modal interaction to extract a concept in the text modality. This concept is aligned with the concept of the corresponding attribute in the visual modality. This is a way to decouple different attributes with the help of the text modality. Therefore, we propose \textbf{Attribute Prototype Learning (APL)} as shown in Figure \ref{fig:method} (left). We define $Query_A$ as the concept of attribute $A$. When an image has attribute $A$, the visual representation of the image will have a high similarity with $Query_A$, otherwise low. A natural idea is to manually generate a query using the textual concept of attribute $A$. However, manual queries often do not achieve the best generalization on downstream datasets. To improve the effect of concept extraction, we use a structure of learnable word vectors as prefixes:

\begin{equation}
Query_A=[V]_1^A[V]_2^A\ldots[V]_n^A[A]
\end{equation}
where $[V]_i^A$ denotes the $i_{th}$ learnable word vector and $[A]$ denotes the text description of attribute $A$. After initialization, we respectively calculate the average similarity of positive, negative, and neutral samples of attribute $A$ to $Query_A$:

\begin{equation}
Center_{A+}=mean_{i \in V_{A+}}(S_{i,Query_A })
\end{equation}
\begin{equation}
Center_{A-}=mean_{i \in V_{A-}}(S_{i,Query_A })
\end{equation}
\begin{equation}
Center_A=mean(Center_{A+},Center_{A-})
\end{equation}
where $A+$ is the set of positive samples of attribute $A$ and $A-$ is the set of negative samples of attribute $A$. After obtaining the above metrics, the optimization loss of $Query_A$ is as follows:

\begin{equation}
\begin{aligned}
Loss_{Query_A}=\\
&MSE\Bigl(tanh(S_{Query_A}-Center_A),label_{A}\Bigl) \\
\end{aligned}
\end{equation}
where $label_{A} \in \{-1,1\}$ is the label of attribute $A$. 

\subsection{Representation Neutralization}
\subsubsection{Analysis of bias in image retrieval}

\begin{table}[t]
\centering
\begin{tabular}{ c|c } 
\hline
Target Attribute & Bias Word \\
\hline
bald, bangs, blond hair,  & smart, stupid, rich, \\ 
fat, double chin, glasses, & poor, happy, sad, \\ 
goatee, gray hair, & noble, humble, nice, \\ 
mustache, sideburns, hat & terrible, kind, evil\\
\hline
\end{tabular}
\caption{\label{Attributes and bias words}The target attribute and bias word used in this work.}
\vspace{-3pt}
\end{table}

\begin{figure}
\centering
\includegraphics[width=0.44\textwidth]{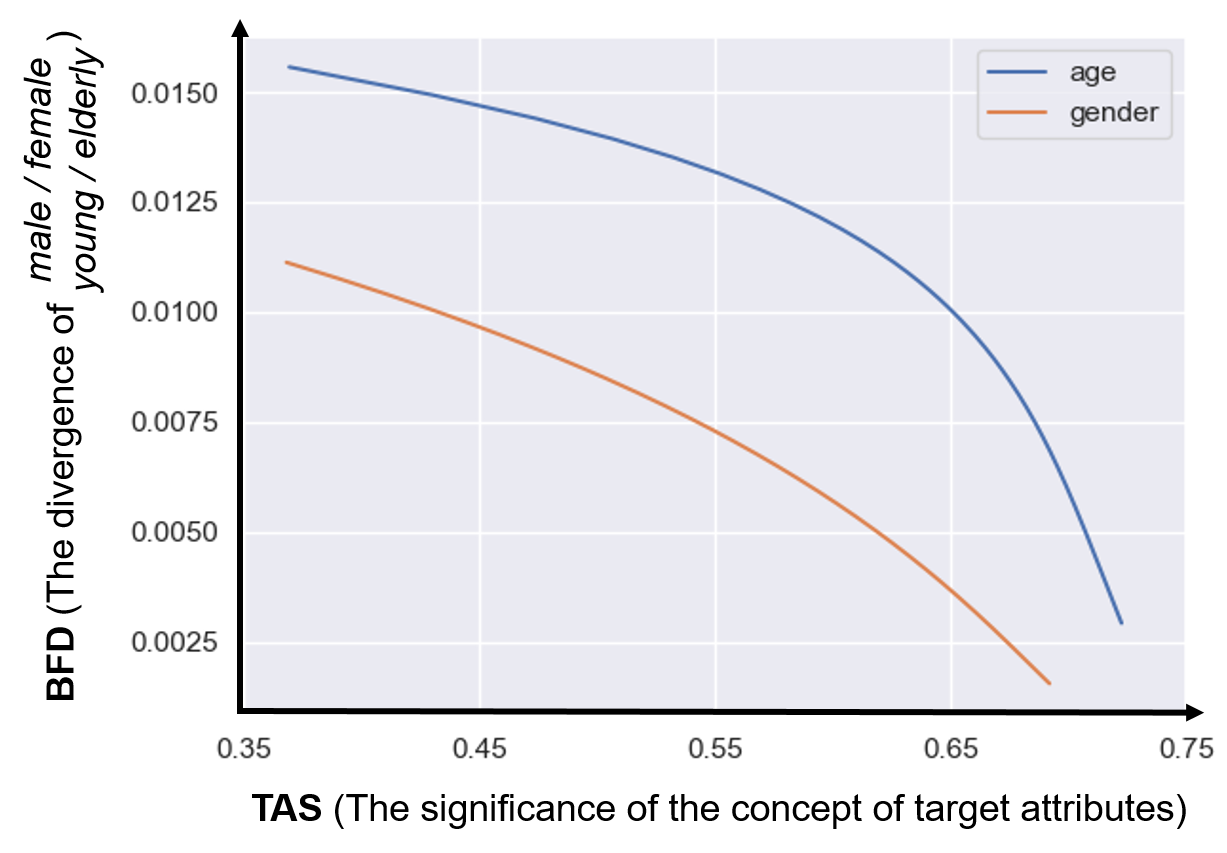}
\caption{\label{fig:TB}The relationship between Target Attribute Significance (TAS) and Bias Feature Divergence (BFD).}
\end{figure}

The bias in image retrieval can be understood from the perspective of attributes. We classify attributes into bias attributes and target attributes based on their contribution to bias. The bias attribute is the attribute that produces the similarity divergence with the retrieval text. The target attribute is the attribute that exists in the sample apart from the bias attributes. Due to differences in the representation of bias attributes, samples with different bias attributes can also differ in their similarity to the same retrieval text. This results in certain groups being more likely to be ranked higher or lower in the retrieval result.

Besides, we find that the significance of the target attributes affects bias. We define the \textbf{Target Attribute Significance (TAS)} and \textbf{Bias Feature Divergence (BFD)} as follows:

\begin{equation}
TAS=S_{Query_j},\ j \in A_T
\end{equation}
\begin{equation}
\begin{aligned}
BFD = MSE\Bigl(&(S_{i,Query_{A_B{}_+}},S_{j,Query_{A_B{}_+}}), \\
&(S_{i,Query_{A_B{}_-}},S_{j,Query_{A_B{}_-}})\Bigl) \\
&where \ i\in A_{B+} \ and \ j \in A_{B-}
\end{aligned}
\end{equation}
where $A_T$ denotes the target attribute, $A_B$ denotes the bias attribute, and $A_B{}_+$ and $A_B{}_-$ denote the positive and negative samples of the bias attribute $A_B$ respectively. TAS measures the significance of samples on target attributes. BFD measures the divergence between the positive and negative samples of the bias attribute. We use gender as an example. We impose adversarial perturbation to control the TAS. And we plot the relationship between TAS and BFD, as shown in Figure \ref{fig:TB}. It can be seen that when target attributes are not significant enough, the samples with different bias attributes have a large divergence in concept representation. As the significance of target attributes increases, the representation divergence shows a decreasing trend.

In summary, we can draw two conclusions. First, divergence in the representation of bias attribute is the root cause of bias. Second, the significance of the target attributes can negatively affect the expression of bias attributes. Therefore, we want to representation neutralization in terms of both target and bias attributes, respectively. For bias attributes, we minimize the representation divergence between positive and negative samples. For target attribute, we improve the significance.

\begin{table*}[t]
\centering
\begin{tabular}{ l|c c c|c c c|c } 
\hline
& & \textit{Gender} & & & \textit{Age} & & \multirow{2}*{\textit{Avg Bias} ($\downarrow$)}\\ 
& CelebA & UTKFace & FairFace & CelebA & UTKFace & FairFace & \\ 
\hline
\textit{Vanilla} & 2.87 & 1.52 & 2.63 & 2.17 & 1.63 & 2.16 & 2.16\\ 
\textit{CLIP-clip} & 2.70 & 2.03 & 3.84 & 1.96 & 1.34 & 2.04 & 2.32(+$7\%$)\\ 
\textit{CD} & 1.61 & 1.46 & 2.59 & 2.62 & 1.65 & 2.18 & 2.02(-$7\%$)\\ 
\textit{AL} & 1.81 & 2.19 & 3.51 & \textbf{1.45} & 1.85 & 1.49 & 2.05(-$5\%$)\\ 
\textit{AA} & 2.68 & 1.64 & 2.51 & 2.18 & 1.58 & 2.13 & 2.12(-$2\%$)\\ 
\hline
\textit{BSCE + RRM} & 2.44 & 1.61 & 2.68 & 1.71 & 1.47 & \textbf{1.48} & 1.90(-$12\%$)\\ 
\textit{APL + RRM} & \textbf{1.59} & \textbf{1.30} & \textbf{1.25} & 1.61 & \textbf{1.12} & 1.54 & \textbf{1.40(-$35\%$)}\\ 
\hline
\end{tabular}
\caption{\label{Debiasing effect}The gender and age $Bias@100$ of different debiasing methods.}
\vspace{-3pt}
\end{table*}

\begin{table*}[t]
\centering
\begin{tabular}{ l |c c c|c c c|c c c|c }
\hline
& \multicolumn{3}{c|}{COCO2017 5K} & \multicolumn{3}{c|}{COCO2014 5K} & \multicolumn{3}{c|}{Flickr 1K} & \multirow{2}*{\textit{Avg Error} ($\downarrow$)}\\
& \textit{R@1} & \textit{R@5} & \textit{R@10} & \textit{R@1} & \textit{R@5} & \textit{R@10} & \textit{R@1} & \textit{R@5} & \textit{R@10} & \\ 
\hline
\textit{Vanilla} & 59.4 & 83.0 & 90.1 & 36.4 & 72.0 & 90.4 & 68.2 & 92.1 & 99.2 & 69.7\\ 
\textit{CLIP-clip} & 55.7 & 79.2 & 87.5 & 32.9 & 64.5 & 88.2 & 62.8 & 89.8 & 98.3 & 79.2(+$14\%$)\\ 
\textit{CD} & 58.0 & 81.5 & 89.0 & 35.1 & 70.3 & 89.4 & 65.1 & 91.3 & 98.6 & 73.9(+$6\%$)\\
\textit{AL} & 44.9 & 70.1 & 79.4 & 26.9 & 58.9 & 80.8 & 56.6 & 86.2 & 96.4 & 99.9(+$43\%$)\\ 
\textit{AA} & 59.5 & 83.2 & 90.2 & 36.5 & 71.8 & 90.6 & 68.5 & 92.4 & 99.3 & 69.3(-$1\%$)\\ 
\hline
\textit{BSCE + RRM} & 59.5 & 82.9 & 90.2 & 36.5 & 17.5 & 90.5 & 68.1 & 92.2 & 99.2 & 69.9($0\%$)\\ 
\textit{APL + RRM} & 58.4 & 81.4 & 89.1 & 36.8 & 70.2 & 89.1 & 66.2 & 92.0 & 98.5 & 73.2(+$5\%$)\\
\hline
\end{tabular}
\caption{\label{Retrial effect}The image retrieval performance of different debiasing methods.}
\vspace{-3pt}
\end{table*}

\subsubsection{Re-Representation Matrix}

In the last subsection, we propose debiasing ideas for the target and bias attributes separately. The aim of both is to reduce the difference in the representation to achieve the \textbf{Representation Neutralization (RN)} of the bias attributes so that CLIP cannot obtain a significant bias concept as shown in Figure \ref{fig:method} (right). We use the \textbf{Re-Representation Matrix (RRM)} which is initialized by a unit matrix with the same dimension as the representation layer of CLIP. In the similarity calculation, we first multiply the visual representation with the RRM. Then, the similarity is calculated with the text representation. This computation process is represented by the following equation:

\begin{equation}
S_{i,L,RRM}=\cos<V_i \cdot RRM,L>
\end{equation}

We propose training constraints of RRM for target and bias attribute, respectively. For the bias attribute, we use comparison learning. We compare the representations of the positive and negative samples after RRM to make them as close as possible. We define the Bias Contrast Loss ($BCL$) on the bias attribute $A$:

\begin{equation}
\begin{aligned}
BCL_A=MSE\Bigl(&(S_{i,Query_{A+},RRM_A},S_{j,Query_{A+},RRM_A}), \\
&(S_{i,Query_{A-},RRM_A},S_{j,Query_{A-},RRM_A})\Bigl) \\
&where \ i\in A+ \ and \ j \in A-
\end{aligned}
\end{equation}

For the target attribute, we want RRM to increase the significance of it. We define the Target Feature Loss ($TFL$):

\begin{equation}
TFL_A=\sum_{i \in V}MSE(S_{i,Query_A,RRM},1)
\end{equation}

Finally the training loss of the RRM corresponding to the bias attribute $A_B$ is as follows:

\begin{equation}
Loss_{RRM_{A_B}}=\lambda \cdot BCL_{A_B}+(1-\lambda)\sum_{i}TFL_{A_{Ti}}
\end{equation}

where $\lambda$ is the hyperparameter in [0, 1] and $A_{Ti}$ is the $i_{th}$ target attribute.

\section{Experiment}

\begin{table*}[t]
\centering
\begin{tabular}{ l|c c c|c c c|c } 
\hline
& & \textit{Gender} & & & \textit{Age} & & \multirow{2}*{\textit{Avg Bias} ($\downarrow$)}\\ 
& CelebA & UTKFace & FairFace & CelebA & UTKFace & FairFace & \\ 
\hline
\textit{BCL} & 1.85 & 1.98 & 2.21 & 2.16 & 1.52 & 2.21 & 1.99\\ 
\textit{TFL} & 2.10 & 2.36 & 2.28 & 1.55 & 1.42 & 1.59 & 1.89\\ 
\textit{BCL+TFL} & 1.88 & 1.98 & 2.21 & 1.58 & 1.20 & 1.70 & 1.76\\ 
\textit{APL+BCL} & 1.90 & 3.02 & 3.35 & 2.15 & 1.59 & 2.20 & 2.37\\ 
\textit{APL+TFL} & 1.92 & 2.18 & 1.77 & \textbf{1.47} & 1.31 & 1.56 & 1.71\\ 
\hline
\textit{ALL} & \textbf{1.59} & \textbf{1.30} & \textbf{1.25} & 1.61 & \textbf{1.12} & \textbf{1.54} & \textbf{1.40}\\ 
\hline
\end{tabular}
\caption{\label{Ablation}We perform ablation experiments for three key settings of FairCLIP: Bias Prototype Learning (APL), Bias Contrast Loss (BCL) and Target Feature Loss (TFL).}
\vspace{-3pt}
\end{table*}

\subsection{Bias Measures}

We note the fairness measure \textbf{Equal Opportunity} proposed in \cite{hardt2016equality}. The unbiased case under this measure is that all samples appear with equal opportunity. We refer to this idea and introduce Equal Opportunity into image retrieval bias measure. A fair retrieval result is that given the $T_B$, the probability of occurrence of a sample with a certain attribute in $R_k$ is close to the probability distribution of this sample group in the whole dataset. Our bias measure is as follows:

\begin{equation}
Bias@k(A,T_B)=|p_{A,T_B,R_K}-p_{A,\tau,R}|
\end{equation}
where $p_{A,T_B,R_K}$ represents the occurrence probability of samples with attribute $A$ in $R_k$ when $T_B$ is used as the retrieval text, and $p_{A,\tau,R}$ represents the occurrence probability of samples with attribute $A$ in the whole image dataset.

\subsection{Dataset, Attribute and Bias Word}

In order to get more annotation of target attributes to participate in the training phase, we choose \textbf{CelebA} \cite{liu2015faceattributes} which has the most labeled face attributes as the training set. There are 40 labeled face attributes in CelebA. For the selection of bias attributes, we choose gender and age. Although race is an important bias attribute, the label of race is very inconsistent in the three face datasets. To avoid ambiguity, we do not use race for bias evaluation. All three face datasets have the same explicit gender labeling. They differ slightly in the bias attribute of age. We divide the sample into young and non-young according to the criteria of CelebA. In UTKFace, age is an integer. We classify samples below 30 years old as young people and the others as non-young people. In FairFace, age is an interval of the length of 10. We select samples aged 20-29 as young and 60-69 as non-young. For the selection of target attributes, we first artificially exclude some attributes with too small scales, such as arched eyebrow and big nose. To ensure that CLIP is sensitive to the selected target attributes, we calculate the representation divergence between positive and negative samples of these target attributes and finally select 11 attributes with relatively significant representation divergence as target attributes. For the selection of bias words, we refer to \cite{berg2022prompt} and choose 12 bias words that may contain human bias. The target attributes and bias words are shown in detail in Table \ref{Attributes and bias words}.

For bias measure datasets, we choose three datasets: \textbf{CelebA}, \textbf{UTKFace} \cite{zhifei2017cvpr} and \textbf{FairFace}\cite{karkkainenfairface}. To improve the robustness of RRM, we use uncropped versions of them. And we select the \textbf{MSCOCO} \cite{lin2014microsoft} and \textbf{Flickr} \cite{young2014image} to evaluate the retrieval performance. The reason why we do not evaluate the retrieval performance using face datasets is that these datasets have limitations that make it difficult to use the labeled target attributes for retrieval. Specifically, the target attributes in the face datasets are generally less significant and difficult for the model to perceive effectively. This leads to poor retrieval results for all methods including Vanilla. In order to evaluate the retrieval performance of different methods more accurately, we refer to the popular retrieval performance evaluation methods.

\subsection{Setup}

We divide CelebA into a training set and a test set. The training set is $30\%$ and the test set is $70\%$. All the debiasing processes are done on the training set. For the selection of CLIP architecture, we use the RN101 architecture, with a 512-dimensional representation layer, a ResNet-101 model for the visual encoder, and a Bert model for the text encoder. In the APL, the initialization of the learnable word vector prefixes is randomly initialized, and the optimizer is SGD. For the choice of hyperparameters, we set the length of the prefixes to 6 and the $\lambda$ to 0.8.

\begin{figure}
\centering
\includegraphics[width=0.44\textwidth]{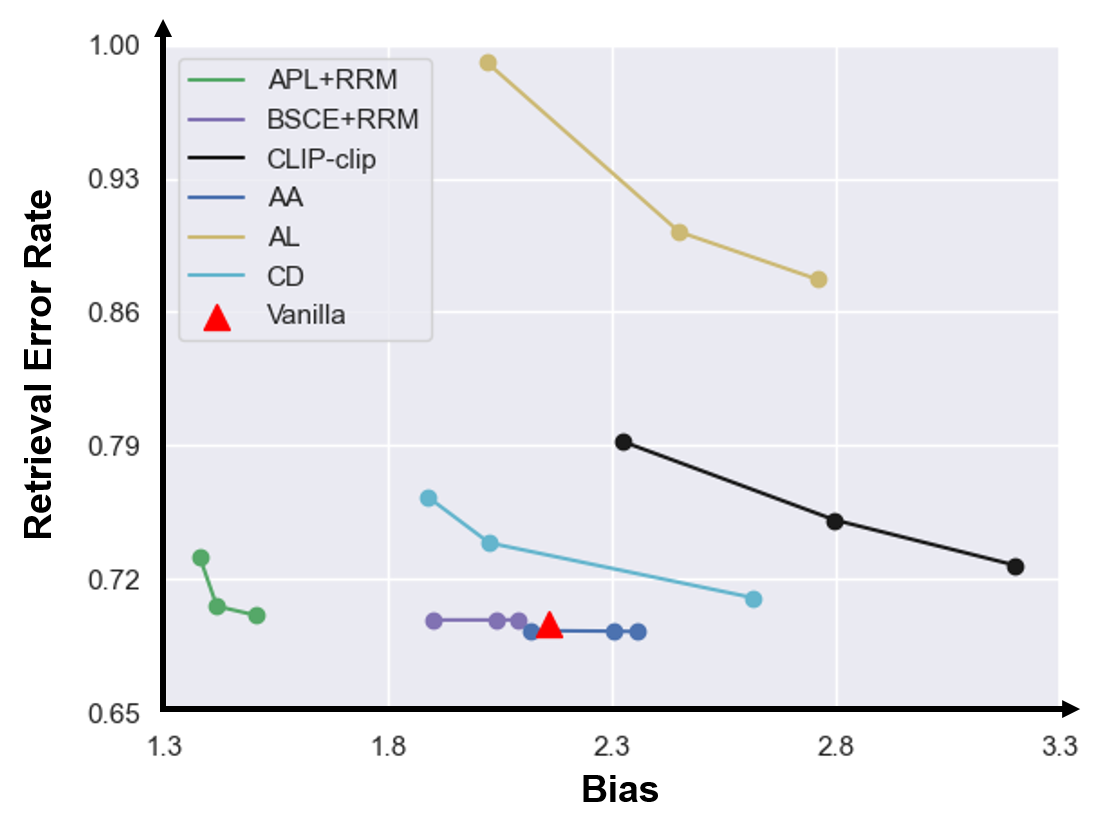}
\caption{\label{fig:RB}We plotted a two-dimensional plot of the debiasing effect and retrieval performance for each debiasing method, where the horizontal coordinate is average $Bias@100$ and the vertical coordinate is the average error rate. We plotted three feature points for each debiasing method by adjusting the parameters.}
\end{figure}

\subsection{Baseline}

To evaluate the debiasing effect of FairCLIP, we select methods \textbf{Adversarial Learning (AL)} \cite{berg2022prompt} and \textbf{CLIP-clip} \cite{wang2021gender} that are proposed to address the CLIP-based image retrieval bias. And we also choose \textbf{Adversarial-Attack (AA)} \cite{zhang2020towards} and \textbf{Contrast Dropout} \cite{meade2021empirical} that are migrated from unimodal fairness work. We also use the \textbf{Bias Space Concept Extraction (BSCE)} proposed in the work \cite{bolukbasi2016man} as the method of concept extraction in the first step to compare with APL.

\begin{figure}
\centering
\includegraphics[width=0.44\textwidth]{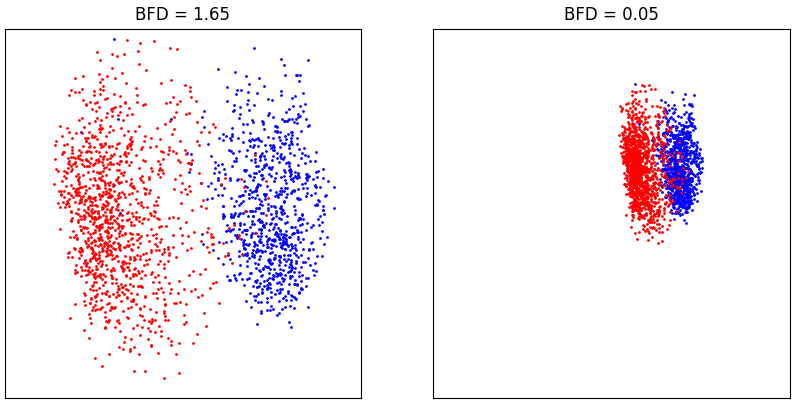}
\caption{\label{fig:PCA}Visualization result for male (blue) and female (red) samples representation. We select the first 2 dimensions of the PCA results for visualization. The left is original data and the right is the debiasing data.}
\end{figure}

\begin{table*}[t]
\centering
\begin{tabular}{ l|c|c c c|c c c}
\hline
& & \multicolumn{6}{c}{\textit{T} = \textit{A photo of a} \textbf{V} \textit{person}} \\
\cline{3-8}
& & \textbf{V} = \textit{happy} & \textbf{V} = \textit{sad} & \textit{Divergence} & \textbf{V} = \textit{nice} & \textbf{V} = \textit{terrible} & \textit{Divergence} \\ 
\hline
\multirow{2}*{\textit{Vanilla}} & \textit{Male} & 57.7 & 42.3& \multirow{2}*{20.1} & 20.1 & 79.9 & \multirow{2}*{13.8} \\ 
 & \textit{Female} & 77.8 & 22.2 &  & 33.9 & 66.1 &  \\
\hline
\multirow{2}*{\textit{FairCLIP}} & \textit{Male} & 91.4 & 8.6& \multirow{2}*{\textbf{2.2}} & 14.0 & 86.0 & \multirow{2}*{\textbf{1.9}} \\ 
 & \textit{Female} & 93.6 & 6.4 &  & 15.9 & 84.1 &  \\
\hline
\end{tabular}
\caption{\label{Discuss}We examine the feasibility of FairCLIP on a classification task. We calculate the average probability of zero-shot classification for male and female samples on given labels.}
\vspace{-3pt}
\end{table*}

\subsection{Result}

\subsubsection{Debiasing effect and retrieval performance}

The result of the debiasing effect is shown in Table \ref{Debiasing effect}. In the Vanilla result, the average total bias was as high as \textbf{2.16}. This means that the probability of a group appearing in the retrieval result of each query consisting of bias word has a 20\% divergence from its probability in the dataset on average. This phenomenon can be particularly serious when certain bias word appears (e.g. as shown at the top of Figure \ref{fig:intro}). As can be seen from the results, CLIP-clip is the only method that does not reduce the bias. This is because the method is designed for open-world datasets. The bias information in the open-world datasets is much weaker than that in the face dataset, which leads to a much lower debiasing effect of CLIP-clip on the face dataset. Compared to the limited debiasing effect of other methods (2\% to 7\%), FairCLIP can achieve 35\%. This benefits from the effective combination of both APL and RRM. When we change the concept extraction to BSCE, the debiasing effect drops dramatically. However, thanks to the superiority of the RRM structure, the debiasing effect is still better than other methods.

In addition to the debiasing effect, we also focus on the impact of debiasing methods on CLIP retrieval performance. We expect a good debiasing method to have a significant impact on the bias attribute and no significant impact on the target attribute. We apply different debiasing methods to the image retrieval. We compute \textit{R@1}, \textit{R@5}, \textit{R@10} and their average error rate. The result of retrieval performance is shown in Table \ref{Retrial effect}. From the result, the AA has almost no effect on retrieval performance. This is due to the fact that the adversarial attack imposes a small perturbation for the representation, and such a perturbation can hardly affect the representation of the target attributes. Except for AA, all other methods showed degraded retrieval performance. The most serious is AL, where the retrieval performance drops by 43\%, and we believe that such a result makes the method almost unfeasible. FairCLIP improved the average error by 5\%. Compared to other methods, this result is second only to AA. Numerically, FairCLIP does not have a significant impact on retrieval performance. Compared with APL, BSCE has less impact on retrieval performance, but this is at the expense of the debiasing effect. 

In order to comprehensively evaluate the superiority of the debiasing methods, we consider both the debiasing effect and the retrieval performance. In Figure \ref{fig:RB}, we plot a two-dimensional scatter plot of bias and retrieval error rate for each debiasing method. We calculated three feature points for each method by parameter adjustment. In the plot, the line corresponding to a good method should be as close to the coordinate axis origin as possible. As can be seen, FairCLIP rides high on the debiasing effect, while the retrieval performance does not receive serious damage.

\subsubsection{Ablation experiment}

We perform an ablation experiment to evaluate the role of each part of FairCLIP. The experiment result is shown in Table \ref{Ablation}. From the first two rows of the table, we can learn that both BCL and TFL can reduce bias. This result can be a good validation of our analysis of bias: both target and bias attributes contribute to bias. And when the two are used in combination, the debiasing effect of the two can be well stacked. Comparing the first three lines with the last three lines, we can learn that APL can effectively improve the debiasing effect. This is because APL is able to extract more precise attribute concepts and thus more targeted when training RRM. Overall, whether it's age bias or gender bias, the best debiasing effect can only be achieved when every part is involved in debiasing. This suggests that each part of FairCLIP has an important role in eliminating bias.

\subsubsection{Visualization}

To visualize the impact of FairCLIP on the visual representation, we visualize the effect of FairCLIP on gender as shown in Figure \ref{fig:PCA}. In the representation space without debiasing, the male and female samples are significantly differentiated and have a large BFD. After using FairCLIP, the BFD is significantly reduced. This means that FairCLIP brings samples of different genders closer together in the direction of gender. When the representations of samples of different genders are relatively close, the differences in representations arising from gender are also reduced. This makes the impact of gender divergence much less when calculating the similarity. This means that regardless of any other task, male and female representations will become closer due to the introduction of FairCLIP.

\section{Discussion}

In this work, we treat the image retrieval task as a fairness issue and propose FairCLIP to eliminate the bias. We choose image retrieval tasks for bias research for two reasons: (1) Many of the applications currently developed applications based on CLIP are image retrieval such as using text to retrial images; (2) Bias in the text can easily manifest in image retrieval. For example, when using ``A photo of a smart person'' to retrieve on CelebA, almost all samples in the first returned 100 samples are male. We calculate the average similarity and find that the male is 0.41 and the female is 0.40. Although the divergence in similarity is low, this gives men a higher probability of being ranked in the top position, resulting in serious bias. Although the FairCLIP targets the image retrieval in this work, it is a general debiasing method in CLIP. FairCLIP achieves the neutralization of representation, which means that the debiasing representation can be effective in other tasks based on CLIP. We examine FairCLIP on a classification task as shown in Table \ref{Discuss}. It can be seen from Vanilla's results that CLIP uses gender as a basis for classification when classifying on labels consisting of a pair of antonyms. This results in a more significant difference between the classification results of males and females. After using FairCLIP, the divergence of classification results is greatly reduced. The result shows that FairCLIP is still effective for the classification task. Accordingly, FairCLIP, as a general debiasing method, is able to debiasing at the representation level, which provides new ideas for eliminating the bias on other tasks related to CLIP.

\section{Conclusion}

In this work, we propose FairCLIP to eliminate the social bias in CLIP-based image retrieval tasks. FairCLIP is divided into two steps: (1) Attribute Prototype Learning; (2) Representation Neutralization. In the first step, to improve the effectiveness of concept extraction, we use the learnable word vector prefixes. In the second step, we analyze the sources of bias and divide the attributes into target and bias attributes. In addition, we propose Re-Representation Matrix (RRM) to achieve debiasing by neutralizing the visual representation. The experiments demonstrate that FairCLIP is a method that has the best compatibility between the debiasing effect and retrieval performance. Although FairCLIP is proposed for image retrieval tasks, FairCLIP achieves the neutralization of representation, which means that the method can be effective in other tasks based on CLIP.

\clearpage
\bibliography{sample}

\end{document}